%% file: fast_paper_camera_ready.tex
\RequirePackage[english=usenglishmax]{hyphsubst}

\documentclass[sigconf,screen]{acmart}
\usepackage{epsfig}                      
\usepackage{graphicx}

\usepackage[inkscapelatex=false]{svg} 

\usepackage{fancyhdr}
\usepackage{flushend}

\usepackage{caption}
\usepackage{subcaption}
\usepackage{listings}
\usepackage{tabularx}
\usepackage{multirow}
\usepackage{makecell} 
\usepackage{enumitem}
\usepackage{verbatim}
\usepackage{graphicx}
\usepackage{algorithm}
\usepackage{algpseudocode}
\usepackage{setspace}
\usepackage[short,nocomma]{optidef}    
\usepackage[normalem]{ulem}

\usepackage{microtype}

\setlist[itemize]{leftmargin=*}

\usepackage[all=normal,    
    wordspacing=tight,     
    floats=tight,
    ]{savetrees}

\setcopyright{rightsretained}
\acmPrice{}
\acmDOI{10.1145/3503222.3507767}
\acmYear{2022}
\copyrightyear{2022}
\acmSubmissionID{asplos22main-p1125-p}
\acmISBN{978-1-4503-9205-1/22/02}
\acmConference[ASPLOS '22]{Proceedings of the 27th ACM International Conference on Architectural Support for Programming Languages and Operating Systems}{February 28 -- March 4, 2022}{Lausanne, Switzerland}
\acmBooktitle{Proceedings of the 27th ACM International Conference on Architectural Support for Programming Languages and Operating Systems (ASPLOS '22), February 28 -- March 4, 2022, Lausanne, Switzerland}

\begin{document}

\title{A Full-Stack Search Technique for Domain Optimized Deep Learning Accelerators}

\author{Dan Zhang}
\email{dazh@google.com}
\affiliation{%
  \institution{Google Brain}
  \city{Mountain View}
  \state{CA}
  \country{USA}
}

\author{Safeen Huda}
\email{safeen@google.com}
\affiliation{%
  \institution{Google}
  \city{Sunnyvale}
  \state{CA}
  \country{USA}
}

\author{Ebrahim Songhori}
\email{esonghori@google.com}
\affiliation{%
  \institution{Google Brain}
  \city{Mountain View}
  \state{CA}
  \country{USA}
}

\author{Kartik Prabhu}
\email{kprabhu7@stanford.edu}
\affiliation{%
  \institution{Stanford University}
  \city{Stanford}
  \state{CA}
  \country{USA}
}

\author{Quoc Le}
\email{qvl@google.com}
\affiliation{%
  \institution{Google Brain}
  \city{Mountain View}
  \state{CA}
  \country{USA}
}

\author{Anna Goldie}
\email{agoldie@google.com}
\affiliation{%
  \institution{Google Brain}
  \city{Mountain View}
  \state{CA}
  \country{USA}
}

\author{Azalia Mirhoseini}
\email{azalia@google.com}
\affiliation{%
  \institution{Google Brain}
  \city{Mountain View}
  \state{CA}
  \country{USA}
}
\renewcommand{\shortauthors}{D. Zhang, S. Huda, E. Songhori, K. Prabhu, Q. Le, A. Goldie, and A. Mirhoseini}

\begin{abstract}
The rapidly-changing deep learning landscape presents a unique opportunity for building inference accelerators optimized for specific datacenter-scale workloads. We propose \textit{Full-stack Accelerator Search Technique} (FAST), a hardware accelerator search framework that defines a broad optimization environment covering key design decisions within the hardware-software stack, including hardware datapath, software scheduling, and compiler passes such as operation fusion and tensor padding.
In this paper, we analyze bottlenecks in state-of-the-art vision and natural language processing (NLP) models, including EfficientNet \cite{efficientnet} and BERT \cite{bert}, and use FAST to design accelerators capable of addressing these bottlenecks.
FAST-generated accelerators optimized for single workloads improve Perf/TDP by 3.7$\times$ on average across all benchmarks compared to TPU-v3.
A FAST-generated accelerator optimized for serving a suite of workloads improves Perf/TDP by 2.4$\times$ on average compared to TPU-v3.
Our return on investment analysis shows that FAST-generated accelerators can potentially be practical for moderate-sized datacenter deployments.\looseness=-1
\end{abstract}



\maketitle

\input{introduction}

\input{relatedwork}
\input{motivation}
\input{body}
\input{results}

\section{Acknowledgements}
The authors would like to thank the anonymous reviewers,
Herman Schmit, Norm Jouppi, Cliff Young, James Laudon, Ed Chi, Priyanka Raina,
Milad Hashemi, Yanqi Zhou, and Kunle Olukotun for their valuable feedback.
\looseness=-1


\interlinepenalty=10000
\bibliographystyle{ACM-Reference-Format}
\bibliography{references}


\end{document}

%% file: introduction.tex
\newcommand{\mytilde}{\raise.17ex\hbox{$\scriptstyle\mathtt{\sim}$}}

\section{Introduction}\label{sec:intro}

The deep learning landscape is constantly evolving. Neural networks nowadays may serve millions or even billions of daily users. Examples include language and image processing models that are used in search engines \cite{BERTinSearch} and social networks \cite{fbAML}. The increasing application of deep learning across different industries suggests that this large-scale adoption trend will only continue to grow. Therefore, we see a potential opportunity for building inference accelerators optimized for specific datacenter-scale workloads.

Enabling automatic hardware optimization requires a search space definition encompassing beyond simple hardware accelerator families such as those used in prior work \cite{has1, has3, magnet, interstellar, miladhas, nahas, apollo, kao2020confuciux, dmazerunner}.
Hardware accelerator architectures can be described in terms of their datapath and schedule, where the datapath comprises the hardware components (compute units, scratchpad memories, connectivity, etc.) on which neural network operations are run, and the schedule comprises the compiler scheduling and hardware control logic that maps these operations onto the datapath.
Common datapath designs use grids of processing elements (PEs), including scalar \cite{interstellar, miladhas, Eyeriss, has1, has3}, vector \cite{magnet, simba, nahas, edgetpu, apollo}, or matrix \cite{tpupaper2017} compute units.
In addition, many compiler optimization passes have major performance impact on production accelerators \cite{NEURIPS2020_9f29450d} and should be included in the search space.

\begin{figure}
\centering
\includesvg[width=0.49\textwidth,keepaspectratio]{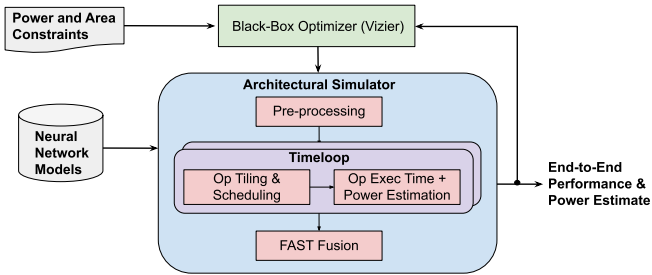}
\vspace{-2.0em}
\caption{Full Stack Accelerator Search Technique overview.} 
\label{fig:archSearch}
\vspace{-1.5em}
\end{figure}

We therefore propose FAST, a Full-stack Accelerator Search Technique that takes one or more neural networks as input, jointly optimizes key decisions within the hardware-software stack including compiler decisions, and outputs an optimized inference accelerator for the input (see Figure \ref{fig:archSearch}).
FAST can optimize for desired objectives such as performance measured in inference queries per second (QPS) or 
performance per Total Cost of Ownership (TCO), a key optimization metric for designing datacenter accelerators \cite{tpupaper2017}.
Unfortunately, TCO is highly sensitive proprietary information;
we instead evaluate performance per Thermal Design Power (TDP), known to highly correlate with Perf/TCO \cite{tpuv4i}.
Our ROI analysis suggests FAST-generated accelerators can be practical for even moderate-sized datacenter deployments, potentially enabling accelerators optimized for single workloads.

\begin{figure}
\centering
\includesvg[width=0.45\textwidth, keepaspectratio]{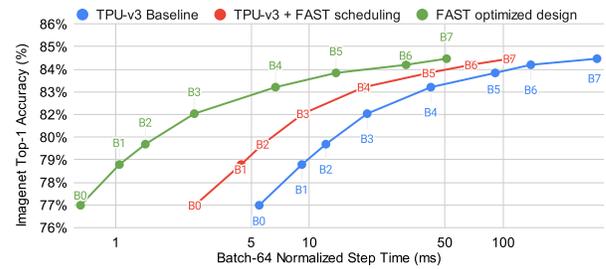}
\vspace{-0.5em}
\caption{FAST-Large (see Table \ref{table:fastDesigns}) inference step time vs. ImageNet top-1 accuracy running the EfficientNet model family. Faster hardware accelerators can run larger, more accurate models with the same latency budget, or significantly reduce inference latency and throughput given a fixed accuracy budget. FAST does not affect model accuracy; quantization can bring further gains but is outside the scope of this paper.} 
\label{fig:accuracyVsPerf}
\vspace{-1.5em}
\end{figure}

Deep learning models are changing rapidly, necessitating similar rapid changes to accelerator architecture designs. State-of-the-art production models, such as EfficientNet \cite{efficientnet} (an image classification model that uses depth-wise separable convolutions) and BERT \cite{bert} (a language model that uses the attention mechanism), introduce a new set of computational and memory bottlenecks that previously did not exist. 
Due to the production impact of these models, we focus on their performance characterization in Section \ref{sec:workloadanalysis}. 
Our analysis demonstrates that EfficientNet has poor operation intensity and efficiently mapping its depth-wise separable convolutions to existing hardware is challenging.
We also describe BERT attention layer inefficiencies, and show that the attention layer becomes a substantial portion of total execution time as the model's sequence length increases.
By designing our search space such that our identified bottlenecks can be properly addressed, we are able to achieve significantly higher Perf/TDP improvements relative to prior work not only on EfficientNet and BERT, but also on older models such as ResNet-50 \cite{resnet50}. 
Figure \ref{fig:accuracyVsPerf} shows the performance of FAST on EfficientNets with scheduling-only updates applied to a fixed hardware configuration, with even larger speedups possible when running full SW/HW co-optimization for each model.

Comprehensively addressing our identified bottlenecks requires a large search space $\mathcal{O}(10^{2300})$ with parameters defining the hardware datapath, software scheduling, and compiler passes such as operation fusion and tensor padding (see Section \ref{sec:fastOverview}).
We extend FAST's datapath template to be an \textit{approximate superset} of existing accelerator families capable of expressing scalar, vector, and matrix processing elements with a versatile memory hierarchy search space.
Our datapath template also includes a TPU-like \textit{vector processing unit} (VPU) \cite{tpuv3} within the PEs which enables efficient execution of a wide range of vector ops, such as exponential and reduction ops, required for workloads such as BERT.
We also discovered that a key limitation of prior work is the inability to generate high-performance accelerators for workloads with low operational intensity, such as EfficientNet. To address this, we devised a flexible and general \textit{integer linear programming} (ILP)-based op fusion technique called \textit{FAST fusion} which can determine the best set of activation and weight tensors to move from DRAM into on-chip scratchpads to maximize overall performance. FAST fusion enables our search tool to unlock significant speedups that are otherwise impossible due to memory bandwidth bottlenecks (see Section \ref{sec:perfBreakdown}) by allowing our search tool to increase scratchpad space to improve fusion efficiency and reduce memory traffic, resulting in high-performance and well-balanced designs (see Table \ref{table:fastDesigns}). 
FAST also considers various optimizations including tensor padding and a new softmax computation approach to the search space, which reduces the memory bottleneck at the expense of additional compute (see Section \ref{sec:softmax}).

A flexible simulator is key to evaluating full-stack accelerator performance for a given neural network. We describe our fast and accurate simulation platform capable of modeling a wide range of hardware datapaths and schedules on unmodified XLA HLO graphs by leveraging Timeloop \cite{timeloop} and addressing its key limitations as discussed in Section \ref{sec:simulator}. Our simulator also contains an analytical power and area model correlated to production designs on an industry sub-10nm process.

\noindent In summary, our contributions are as follows:
\begin{itemize}[leftmargin=*]
\item We propose FAST, an automated framework for jointly optimizing hardware datapath, software schedule, and compiler passes, with a combined search space of up to $\mathcal{O}(10^{2300})$ to design optimized inference accelerators for one or a set of input neural networks. 
\item We perform detailed performance characterization of state-of-the-art ML models, identify their bottlenecks, and propose several optimizations to address these bottlenecks.
\item We propose a FAST fusion, a novel and flexible ILP-based op fusion technique, enabling FAST to fully address memory bottlenecks in low operational intensity workloads.
\item We analyze the relationship between ROI, number of deployed hardware, and the Perf/TDP of accelerators to provide guidelines on exploring trade-offs between specialization and performance in future accelerators. Our ROI analysis demonstrates that FAST-generated inference accelerators can potentially be practical for even moderate-sized datacenter deployments (Section \ref{sec:results}).
\item We evaluate FAST on a comprehensive set of models, including the EfficientNet family \cite{efficientnet}, BERT \cite{bert}, ResNet50v2 \cite{resnet50} and production OCR workloads \cite{vssOcr}.
\item FAST's custom designs demonstrate an average of 3.7$\times$ Perf/TDP improvement across all the benchmarks when compared against TPU-v3, including a 6.4$\times$ and 2.7$\times$ improvement for EfficientNet and BERT respectively.
\item FAST's general-purpose designs optimized for serving a set of important vision and NLP benchmarks show an average of 2.4$\times$ improvement in Perf/TDP vs. TPU-v3. 

\end{itemize}

%% file: relatedwork.tex
\section{Related Work}

\noindent \textbf{Accelerator design space exploration:} Hardware ML accelerators can be described in terms of their hardware datapath and software schedule. Datapath designs often use grids of uniform processing elements comprised of scalar \cite{interstellar, miladhas, Eyeriss, has1, has3, dmazerunner}, vector \cite{magnet, simba, nahas, edgetpu, apollo}, or matrix \cite{tpupaper2017} compute units. 
Previous work focuses on optimizing families of accelerator designs with scalar or vector PEs and fixed memory hierarchies by mutating datapath hyperparameters, such as the number of PEs and buffer capacities \cite{has1, has3, magnet, interstellar, miladhas, nahas, apollo, kao2020confuciux, dmazerunner}, as well as the mapping of convolutions onto the datapath \cite{interstellar, miladhas, magnet, kao2020confuciux, has3, dmazerunner, analytical, mindmappings, gamma}.

As described in Section \ref{sec:params}, our datapath template is designed to be an approximate superset of popular designs capable of expressing scalar, vector, and matrix processing elements with varying memory hierarchies beyond variations on accelerator families.
Our PEs containing both vector and systolic units share similarity to heterogeneous PE designs such as Plasticine \cite{plasticine}. 
Our search space also includes scheduling and other compiler optimizations, such as op fusion and tensor padding, enabling us to cover a much broader set of architectures. 
We also optimize for state-of-the-art models including EfficientNet and BERT, and demonstrate that our large co-optimization space allows for significant improvements over existing datacenter accelerators.

A flexible scheduler is key to evaluating accelerator performance for a given neural network. Timeloop \cite{timeloop} and MAESTRO \cite{maestro} use random search to optimize accelerator schedules given a datapath and layer definition. However, they only evaluate single layers and only consider convolution operation performance, thereby limiting utility for end-to-end performance evaluation and optimization (e.g., operator fusion, parameter prefetching). Compared to Timeloop, the MAESTRO datapath design space is more restrictive, assuming only NVDLA accelerator variants with private L1 and global L2 scratchpads, and can only fuse ReLU and pooling operations. Interstellar \cite{interstellar} uses Halide \cite{halide}, a domain-specific programming language, to generate and analyze inference accelerators. Although Interstellar can perform many blocking and spatial optimizations, its datapath search space is limited to grids of scalar PEs and reduction trees with global buffers. dMazeRunner \cite{dmazerunner} optimizes only convolution and FC layers with a pruned search of the schedule space. ZigZag \cite{zigzag} has a flexible datapath design space, and can perform heuristic-based schedule search targeting a scalar PE architecture, but does not perform fusion.
We use Timeloop while addressing its limitations in our simulator, as described in Section \ref{sec:simulator}.
Our comprehensive search space led to significantly larger speedups compared to prior work; for example, MAGNet’s reported best result \cite{magnet} only improved Perf/W by 1.75x (43\% energy reduction) compared to our best reported result of 6x Perf/TDP.
To the best of our knowledge, FAST is also the first to optimize designs across multiple workloads.

Recent work such as ASIC Clouds \cite{asicCloudsDSE, asicClouds} has used design space exploration to optimize directly for datacenter total cost of ownership in the context of bitcoin mining, video transcoding, and machine learning accelerators. FAST extends this by considering return-on-investment (ROI) and using ROI to demonstrate production feasibility of FAST-generated designs.

\noindent \textbf{Accelerator search on Reconfigurable Hardware:} Several recent efforts have targeted the acceleration of neural networks on reconfigurable hardware including FPGAs and spatial arrays, which unlike ASICs enable flexible hardware reconfiguration. These prior works primarily focused on automation tools and design space exploration for one particular neural network \cite{has2, FPGA1, FPGA2, FPGA3, FPGA4, fpga7, fpga8, fpga9, fpga10, fpga11, fpga12, fpga13, plasticine}. However, the flexibility of FPGAs comes at the cost of reduced performance and higher energy consumption \cite{fpgaGap}. Unlike prior work, our framework enables the exploration of a broad range of datapaths, schedule, and fusion. We believe it would be simple to adapt our work to target reconfigurable hardware.

\noindent \textbf{Co-optimization of neural networks and hardware:} More recently, co-optimizing neural networks and accelerators has gained significant attention \cite{FPGA4, FPGA5, FPGA6, DBLP:journals/corr/abs-1907-04650, codesign-yujun2019, nahas-edd2020, nahas-dac2020, nahas}. The design space contains both the neural network architecture and hardware components, while jointly optimizing for both accuracy and performance. While our framework does not currently allow modifications to the model architecture, it would be straightforward to extend. However, even without model changes, FAST already delivers significantly higher performance than previous work through the larger search space covering datapath, schedule, and fusion.

\noindent \textbf{Operation fusion:}
We also developed FAST fusion, an efficient ILP-based multi-layer fusion technique for inference which significantly improves memory bandwidth usage efficiency and thus inference execution time.
Most production compilers such as cuDNN \cite{cudnn} can only fuse simple pre-defined templates such as Conv2D+Bias+Add.
Although XLA \cite{xla} can create large fusions, each XLA-generated HLO fusion region contains at most one matrix operation (Conv2D, einsum, matmul, etc).
There is also a growing body of work on more elaborate fusion \cite{has2, VLDB18-fusion, NEURIPS2020_9f29450d,fusion-cong2019, roesch2019relay-fusion, mangpo-fusion-2020, zhaofusion}. \cite{learntofuse} presents an RL-based approach for op fusion for training. \cite{memorymanagement} designs a framework for efficiently utilizing FPGA on-chip memory.
FAST fusion is a secondary pass that fuses existing XLA-generated HLO fusion regions by assigning intermediate tensors from DRAM to on-chip SRAM. 
Compared to prior approaches, FAST fusion considers weight tensor pinning as part of the fusion problem, and uses ILP to directly minimize total execution time as modeled through simulation rather than indirect metrics such as total memory accesses.

%% file: motivation.tex
\section{Background}\label{sec:background}


\subsection{Mapping Convolutions onto Accelerators}

Since convolutions dominate the overall runtime in convolutional neural networks (CNNs), considerable effort has been expended on software \cite{fastconvs} and hardware \cite{ShiDianNao, Eyeriss} acceleration of these operations.
A standard Conv2D can be represented as a 7-dimensional nested loop 
over batch size (\textit{B}), output tensor height and width (\textit{OH}, \textit{OW}), number of input 
and output features (\textit{IF}, \textit{OF}), and kernel height and width (\textit{KH}, \textit{KW}). 
Since these loop iterations are commutative, compilers can freely modify loop traversal order,
allowing for arbitrary transformations in tensor layout format, loop blocking, and spatial vectorization \cite{timeloop, maestro}.
Recent work has exploited these properties to build efficient high-performance accelerators
\cite{kao2020confuciux, interstellar, miladhas}.
\looseness=-1

Systolic arrays combine parallel operations with local communication, making them well-suited for matrix computations \cite{systolicarray}.
To multiply two matrices, one matrix is latched into internal registers, while the other is streamed through the array. 
Double-buffering is typically employed to mask the latency of latching a new set of parameters into the systolic array \cite{tpupaper2017}
Accelerators such as Google's TPU family \cite{tpuv3} exploit the dense compute enabled by systolic arrays to accelerate training and inference.
Under a \textit{weight stationary} mapping \cite{brainwave}, the systolic array will not be fully utilized unless \textit{IF}, \textit{OF}, and \textit{B} are multiples of the dimensions of the systolic array.
Alternative mappings, such as \textit{output stationary} and \textit{row stationary} \cite{Eyeriss}, may achieve higher utilization by selecting alternative dimensions to be spatially unrolled, but are still limited by dimensional constraints. 
Therefore, although larger systolic arrays improve area-density and power-efficiency per FLOP, they tend to have lower utilization.
Each workload has different problem shapes, thus having different optimal systolic array dimensions which can be found through FAST. 


\subsection{EfficientNet Overview}

Convolutional neural networks (CNNs) are often over-parameterized \cite{squeezenet, pruning}. 
A popular method for reducing model size and compute cost is replacing Conv2D with a \textit{depthwise-separable convolution}: a depthwise convolution combined with a 1x1 point-wise convolution \cite{xception, mnasnet, mobilenetv2}. 
For example, a 3x3 depthwise-separable convolution uses 8-9x less compute than a standard Conv2D with only a slight reduction in accuracy \cite{mobilenet}.
EfficientNet \cite{efficientnet}, a CNN based on \textit{inverted residual} (MBConv) \cite{mobilenetv2} blocks, demonstrated that depthwise-separable convolutions were viable outside of compute and storage-constrained settings.
However, depthwise-separable convolutions do not map well onto TPUs due to poor systolic array utilization and operational intensity.
Depthwise convolutions allow significant parameter and compute reduction by reducing kernel filter depth (\textit{IF}) to 1, but number of FLOPS is not an accurate proxy for performance on state-of-the-art accelerators such as Google TPUs or NVIDIA GPUs \cite{proxylessnas}. Common mappings unfortunately depend on large \textit{IF} for good utilization.
For example, assuming a depthwise convolution with a 3x3 kernel, maximum utilization for a 128x128 systolic array is only $KH*KW=9$ out of 128.
To address this, EfficientNet-X replaces some depthwise-separable convolutions with Conv2Ds to improve accuracy and latency \cite{efficientnetx}.
However, the poor performance of depthwise convolutions remains a challenge. 
As shown in Table \ref{table:fastDesigns}, FAST optimized for EfficientNet automatically generates hardware with smaller systolic arrays, improved scheduling, and reduced memory bottlenecks, enabling EfficientNet inference at high efficiency.

\subsection{BERT Overview}

Transformer-based models outperform traditional recurrent neural networks (RNNs) and long short-term memory networks (LSTMs) on natural language processing tasks by replacing sequential computation with the self-attention mechanism \cite{transformer}. BERT \cite{bert} is a Transformer-based model that achieves state-of-the-art results on both word-level and sentence-level tasks, and is the inspiration for a number of NLP models, including XLNet\cite{xlnet},  GPT-2 \cite{gpt2}, GPT-3 \cite{gpt3}, ALBERT \cite{albert}, and RoBERTa \cite{roberta}. 
\looseness=-1

BERT is composed of multiple transformer encoder layers, where each layer consists of a self-attention layer, softmax operation, feed-forward layer, residual connection, and layer normalization. An important hyperparameter is the sequence length, which controls the size of the input token sequence. Increasing sequence length generally improves task accuracy at the cost of computation, with some operations scaling more efficiently than others, as discussed in Section \ref{sec:bertPerf}. Although most operations are matrix-matrix multiplications, vector operations such as softmax and layernorm cannot be ignored.
\looseness=-1

\section{Workload Performance Analysis}\label{sec:workloadanalysis}
In the following section, we analyze various contributing factors to EfficientNet and BERT performance on TPU-v3. We first characterize EfficientNet and BERT in terms of operational intensity and discuss the impact of op fusion. We then analyze the implications of TPU-v3 architecture and compute scheduling strategy on EfficientNet. Finally, we examine BERT performance as a function of sequence length. These characterizations motivated us to build a comprehensive hardware and software search space for FAST able to deliver significant performance improvements. 

\subsection{Operational Intensity and Op Fusion}
\label{sec:opIntensityAndFusion}

\begin{table} 
\centering
\resizebox{0.34\textwidth}{!}{%
\begin{tabular}{| l | l | l |}
  \hline
  \textbf{Model} & \textbf{Max Working Set} & \textbf{Weights}\\
  \hline
  EfficientNet-B0 & 2.87 MiB & 12.7 MiB\\
  \hline
  EfficientNet-B1 & 3.3 MiB & 22.1 MiB\\
  \hline
  EfficientNet-B2 & 3.9 MiB & 26.1 MiB\\
  \hline
  EfficientNet-B3 & 5.1 MiB & 36.8 MiB\\
  \hline
  EfficientNet-B4 & 12.4 MiB & 61.4 MiB\\
  \hline
  EfficientNet-B5 & 17.8 MiB & 101 MiB\\
  \hline
  EfficientNet-B6 & 31.9 MiB & 146 MiB\\
  \hline
  EfficientNet-B7 & 41.2 MiB & 231 MiB\\
  \hline
\end{tabular}}
\caption{EfficientNet on-chip storage requirements (bfloat16). Working set sizes are shown for the op with the largest memory footprint at batch size 1. The storage requirements of larger EfficientNets exceed on-chip memory capacity, requiring more advanced op fusion techniques.}
\label{table:efficientNetSizes}
\vspace*{-2.8em}
\end{table}

\begin{figure*}[]
\centering
\vspace{-0.2em}
\includesvg[width=0.99\textwidth, keepaspectratio]{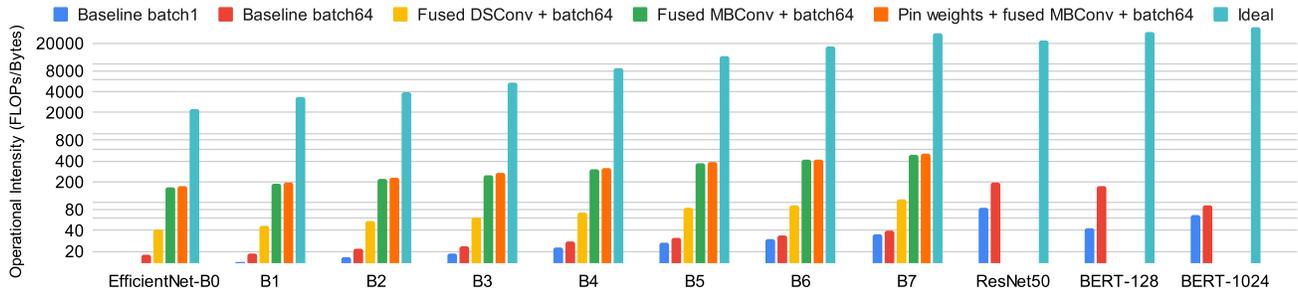}
\vspace{-1.3em}
\caption{The impact of op fusion on operational intensity. Models with op intensity below 200 are memory bandwidth-bottlenecked on current accelerators. BERT and ResNet-50 do not contain depthwise-separable convolution (DSConv) or inverted residual (MBConv) blocks. Increasing batch size is effective for ResNet-50 and BERT-seq128, but not for EfficientNet and BERT-seq1024. Supporting future accelerators with op intensity over 400 requires more advanced fusion techniques.} 
\label{fig:opint}
\vspace{-0.5em}
\end{figure*}

ML model graphs are executed on accelerators as a series of kernels, or \textit{operations}, where each op reads its inputs from device memory (DRAM), 
transfers these inputs to on-chip memory, performs the computation, and writes the output back to DRAM. 
This results in unnecessary DRAM reads and writes for intermediate values which are usually performed in parallel with computation, but may cause slowdowns with insufficient bandwidth.
To determine if a model is compute or memory bandwidth-bound, one can calculate a model's \textit{operational intensity}, defined as the ratio of compute operations (in FLOPS) to DRAM accesses (in bytes).
For example, a TPU-v3 chip supports 123 TFLOPS/s of bfloat16 compute and 900GB/s memory bandwidth \cite{tpuv3}.
Therefore, a model that can operate at full compute utilization must have an operational intensity of at least 137 FLOPS/B to avoid becoming memory-bound. 
Note that it is cheaper to scale compute performance than memory bandwidth due to the \textit{memory wall} \cite{memwall}. 
The latest NVIDIA A100 GPU supports 312 TFLOPS bfloat16 with 1.5TB/s bandwidth \cite{a100}, requiring an operational intensity of 208 FLOPS/B to prevent bandwidth bottlenecks.

Compilers such as TensorFlow XLA \cite{xla} mitigate this issue with \textit{operation fusion}, merging multiple ops into one large op to avoid DRAM accesses of intermediate results, resulting in greater operational intensity and improved performance \cite{tpuv3}.
Most prior work has focused on training, where intermediate results must be preserved for the backwards pass \cite{learntofuse, fusionstitching}.
In this work, we focus on inference, which does not require a backwards pass, meaning that intermediate results may be immediately discarded after use. 

Figure \ref{fig:opint} shows that EfficientNet has low operational intensity due to its heavy use of depthwise-separable convolutions.
Without op fusion, EfficientNet operational intensity ranges from 13 to 35 FLOPS/B, far below the level required to run without memory bottlenecks on TPU-v3 or A100. Using batching to amortize weight accesses across multiple inferences is effective for ResNet-50 and moderately effective for BERT, but not for EfficientNet due to its lower parameter count. As such, these workloads present a significant challenge to architects, since provisioning greater memory bandwidth can result in exorbitant incremental costs. Current fusion approaches are based on templates comprising specific compiler-defined sequences of ops \cite{cudnn}; we consider hypothetical depthwise-separable and MBConv fusion templates. By fusing entire MBConv blocks, we are able to achieve an operational intensity greater than 200 FLOPS/B. However, writing custom block fusion templates is not scalable. There is also considerable operational intensity headroom remaining, as shown by the ideal case in which all model weights are pinned \cite{brainwave} and only the input and final output results require off-chip accesses. 
These insights motivate \textit{FAST fusion}, an fusion technique for inference capable of fusing arbitrary sequences of ops as described in Section \ref{sec:fusion}, addressing the memory bottleneck.

Aggressive op fusion and weight pinning can come at the cost of significant on-chip storage capacity, as shown in Table \ref{table:efficientNetSizes}. 
An op's working set size is the size of its input activations and outputs, and a model's working set size is the working set size of its largest op.
Since working sets scale linearly with batch size, fusion tends to perform better at smaller batch sizes since more tensors will fit into SRAM. 
However, larger batch sizes can improve systolic array utilization, resulting in higher overall performance.
Determining the best resource allocation between compute and memory depends on the specific operational intensity, memory footprint, and batch size for a target workload.
FAST can automatically explore this space through datapath, scheduling, and fusion co-optimization.

\subsection{EfficientNet Resource Utilization}
\label{sec:efficientnetPerf}
\begin{figure}[]
\centering
\vspace{-0.3em}
\includesvg[width=0.47\textwidth, keepaspectratio]{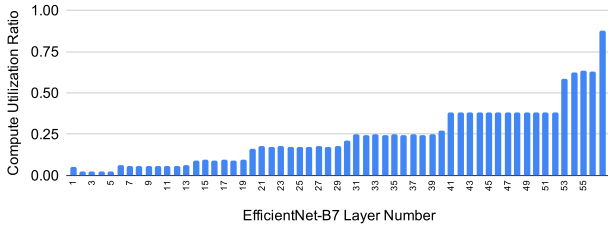}
\vspace{-1.5em}
\caption{EfficientNet-B7 per-layer performance as a fraction of peak FLOPS on TPU-v3. Earlier layers have low utilization due to having few channels. A good utilization ratio should exceed 0.7. Smaller EfficientNets have worse utilization due to having fewer channels.} 
\label{fig:perlayerperf}
\vspace{-1.5em}
\end{figure}

We profiled EfficientNet-B7 performance on TPU-v3. 
Figure \ref{fig:perlayerperf} shows the performance of each MBConv block  as a fraction of peak TPU-v3 compute (FLOPS). 
Initial layers have poor utilization, with utilization improving as the number of input/output channels increases. 
Overall TPU-v3 utilization on EfficientNet-B7 is only 14.8\%, suggesting a potential 6.75x performance upside with an improved datapath and scheduler with similar peak FLOPS that can reach full utilization.

\begin{table} 
\centering
\vspace*{-0.0em}
\resizebox{0.45\textwidth}{!}{%
\begin{tabular}{| l | l | l |}
  \hline
  \textbf{Op Type} & \textbf{FLOP Percentage} & \textbf{Runtime Percentage}\\
  \hline
  DepthwiseConv2dNative & 5.00\% & 65.30\% \\
  \hline
  Conv2D & 94.67\% & 34.20\% \\
  \hline
  Other & 0.33\% & 0.50\% \\
  \hline
\end{tabular}}
\vspace{0.3em}
\caption{EfficientNet-B7 per-op performance as a fraction of total execution time on TPU-v3. Depthwise convolutions consume the majority of execution time, due to their poor mapping efficiency on TPU-v3.}
\label{table:perOpPerf}
\vspace*{-1.5em}
\end{table}

To identify the cause of low average utilization, we examined EfficientNet-B7 operation performance as a fraction of total execution time on TPU-v3 as shown in Table \ref{table:perOpPerf}.
The culprit is clear: depthwise convolutions comprise the majority of overall runtime, but only utilize a small fraction of total compute.
An accelerator design that balanced depthwise convolution and regular convolution performance would therefore see significant speedups on EfficientNet.
We discuss how this can automatically be achieved through FAST. 
\looseness=-1

\subsection{BERT Resource Utilization}
\label{sec:bertPerf}

\begin{figure}
\centering
\vspace{-1.0em}
\includesvg[width=0.47\textwidth, keepaspectratio]{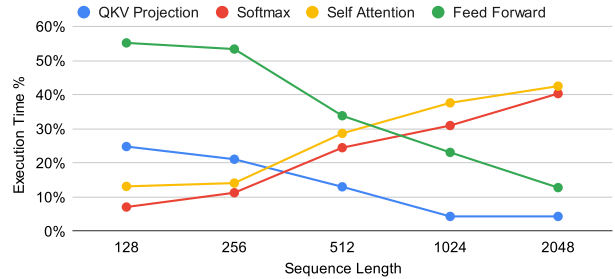}
\vspace{-0.5em}
\caption{BERT per-op performance on TPU-v3. Softmax and self-attention ops, which run inefficiently on TPU-v3, dominate execution time at longer sequence lengths.} 
\label{fig:bertseq}
\vspace{-1.5em}
\end{figure}

We profiled BERT-Base \cite{bert} performance on TPU-v3 with default hyperparameters, sweeping sequence length from 128 to 2048. 
Since each BERT layer is identical, we broke a single layer into its subcomponents: Query/Key/Value matrix projection, softmax, self-attention, and feed-forward.
The QKV projection and feed-forward ops already run efficiently on TPU-v3, at 65\% and 75\% compute utilization respectively. 
Softmax is comprised of vector operations and thus must entirely execute on the TPU-v3 vector unit instead of its systolic array, so has exceptionally low compute utilization as defined as a fraction of peak throughput of less than 1\%. 
Self-attention performs an \textit{activation $\times$ activation} matrix multiply instead of \textit{activation $\times$ weight} such that the cost of latching a matrix into the systolic array cannot be fully amortized over the batch dimension, resulting in lower utilization.
\looseness=-1

As shown in Figure \ref{fig:bertseq}, at low sequence lengths the efficient QKV projection and feed-forward ops dominate execution time, resulting in overall efficient execution.
However, QKV projection and feed-forward computationally scale linearly with sequence length, whereas softmax and self-attention scale quadratically $\mathcal{O}(N^{2})$.
Therefore, the inefficient softmax and self-attention ops dominate execution time at longer sequence lengths, resulting in poor overall performance.
Self-attention performance can be addressed with smaller systolic arrays automatically discovered through FAST.
We discuss techniques for addressing softmax performance in Section \ref{sec:softmax}.

%% file: body.tex
\section{Full-Stack Accelerator Search}


FAST is a full-stack accelerator search technique for automatically designing custom accelerators optimized for a given set of ML workloads and subject to constraints as shown in Figure \ref{fig:archSearch}. We first consider if such techniques are practical, before describing the framework in detail in the following sections.

\subsection{The Economics of Specialized Accelerators}
\label{sec:roi}

It is well-known that increased hardware specialization improves performance \cite{catapult, tpupaper2017, cuda, minnow}.
However, given that specialized accelerators target fewer workloads than general accelerators and building custom chips is expensive, it is less clear whether such specialization is economically viable.
We analyze this question by examining \textit{Return on Investment (ROI)}, a common profitability metric measuring an investment's return relative to its cost \cite{roi}; an ROI exceeding 1 is profitable. 
Companies typically aim to reach a predefined ROI threshold for their projects.
A proper ROI calculation should be based on a company's specific circumstances.
The following analysis is hypothetical, is based only on publicly-available data, and is intended to be used for illustrative purposes. 

A simple method for estimating investment return is based on the savings from deploying a more cost-efficient accelerator relative to the baseline, typically either the currently-deployed accelerator or a next-generation design under consideration.
Suppose we design an accelerator optimized for EfficientNet with a higher Perf/TCO relative to the baseline, and plan to offload all datacenter traffic currently running EfficientNet onto this new accelerator, i.e., aggregate QPS served by the new accelerator will be the same as the current accelerator. 
The ROI for this accelerator can be estimated as:
\begin{align}
  TCO_{old} = C_{cap}(n)+t_{D} \cdot C_{op}(n) \\
  ROI = \frac{TCO_{old}\cdot(S-1)}{(t_{design}\cdot C_{eng}+C_{mask}+C_{IP})\cdot S}
\end{align}

where $C_{cap}(n)$ and $C_{op}(n)$ are the capital and operational costs, respectively, to deploy $n$ accelerators, $t_{D}$ is the accelerator deployment lifetime in years, $S$ is the Perf/TCO improvement relative to the baseline accelerator, $t_{design}$ is the aggregate engineering-years to design the accelerator and its system software, $C_{eng}$ is the corporate cost per engineer per year including compensation, benefits, and all overhead, $C_{mask}$ is the wafer mask cost, and $C_{IP}$ is the IP licensing cost such as the DRAM PHY. All pricing and power numbers for the baseline and new accelerator should include shared system server components, including the fractional host machine, networking, and rack infrastructure amortized between several accelerators.

\begin{figure}[]
\centering
\includesvg[width=0.45\textwidth,keepaspectratio]{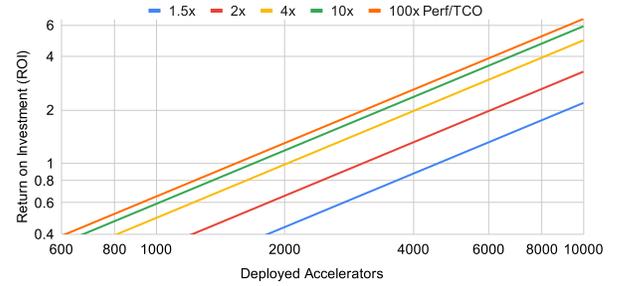}
\vspace{-0.7em}
\caption{Accelerator Return on Investment (ROI) vs. deployment volume for hypothetical specialized accelerators with improved Perf/TCO relative to NVIDIA A100 baseline. An ROI exceeding 1 is profitable. Modern datacenters typically deploy thousands of accelerators, enabling initial engineering and manufacturing costs to be amortized.
\looseness=-1} 
\label{fig:roi}
\vspace{-2.0em}
\end{figure}

Our example ROI calculation assumes a NVIDIA DGX A100 320GB platform baseline containing 8x A100 accelerators with a manufacturer's suggested price (MSRP) of \$199,000 \cite{dgxCost}. 
We assume the May 2021 average price of electricity for the US Commercial sector (\$0.1084/kWh) from the US Energy Information Administration \cite{eiaCost}, an accelerator deployment lifetime of 3 years \cite{supermicro}, the cost per engineer based on the reported median total compensation for a SWE working in the San Francisco bay area (\$240,000) \cite{levelsfyi} with a 65\% salary overhead \cite{asicClouds}, and all other values from previous work \cite{asicClouds}. 
Since our experimental results assume a sub-10nm process technology, we extrapolate wafer mask and PHY IP costs using exponential scaling as observed in \cite{asicClouds}.

Estimating aggregate engineering years is more difficult, since it varies widely on a project methodology and company basis.
Modern chip design has significantly reduced the engineering effort required for custom accelerators using techniques including HLS \cite{argos, simbaHls} and agile design methodologies \cite{riscvAgile}.
Custom accelerators can further reduce design time by leveraging existing hardware and software infrastructure.
Simba \cite{simba} was built by 5-10 engineers in 20 months (12.5 engineer-years) to go from architecture to tape-out \cite{simbaHls}.
The Tesla Full Self Driving (FSD) SoC was built by 100 engineers \cite{muskInterview} in 14 months \cite{teslaFsd} (117 engineer-years). 
Since Simba is a research test chip, and FSD is a full SoC containing a custom ML inference accelerator, we average the two designs to estimate the effort for a dedicated ML inference accelerator (65 engineer-years). 

To approximate accelerator deployment volumes, a naive approach may be to divide a workload's QPS by the accelerator's throughput to estimate the total number of accelerators required to serve a certain amount of traffic.
However, datacenters in practice are heavily over-provisioned to lower response latency and account for issues including traffic spikes, reliability, and projected future user growth \cite{warehousescale}. 
These provisioning calculations are highly confidential; we therefore looked at public examples in industry.
Microsoft deployed 1,632 Catapult FPGAs for a medium-scale pilot study to accelerate a portion of the Bing search ranking pipeline \cite{catapult}; currently, more than a million Catapult FPGAs are deployed in Microsoft datacenters \cite{putnamKeynote}.
Google has deployed tens of thousands of servers with video transcoding accelerators \cite{argos}.
Facebook trains its Facer model on thousands of servers \cite{fbAML}.
Large language models such as Meena \cite{meena}, GShard \cite{gshard}, Switch Transformer \cite{switchTransformer}, and GPT-3 \cite{gpt3} take 1024 to 10,000 accelerators to train \cite{carbonLanguageModels}; a McKinsey study showed that datacenter inference demand typically exceeds training \cite{mckinsey}.

Figure \ref{fig:roi} shows ROI as a function of the number of deployed accelerators, assuming hypothetical specialized accelerators capable of improving Perf/TCO from 1.5x to 100x relative to the NVIDIA DGX A100 baseline.
There are several key takeaways. 
Firstly, a large deployment volume is the most important factor: all accelerators with positive Perf/TCO relative to the baseline become ROI-positive with sufficient volume. 
Secondly, there are diminishing returns to improving Perf/TCO under our strict definition of ROI. For example, deploying 8000 accelerators with 1.5x Perf/TCO has higher ROI than deploying 2000 accelerators with 100x Perf/TCO.

However, our ROI calculation is conservative since it only captures the returns from switching to a more cost-effective platform. Improving inference latency can increase revenue \cite{costlo}; a 500ms delay in the Bing search engine reduced revenue per user by 1.2\% \cite{lowlatency}.
A custom accelerator may also enable larger models that are currently infeasible for deployment on current platforms.
Finally, a new accelerator deployment should not just replace the current accelerator baseline, but account for future application growth; Facebook DL inference server demand increased by 3.5x over less than two years \cite{fbDL}.
Therefore, even accelerators that simply break-even in ROI while enabling significantly lower latency or larger models can potentially be economically viable with sufficient justification. 


\subsection{Problem Definition}

Our objective is to find an optimized set of hyperparameters \textit{h} for the hardware datapath, scheduler, and op fusion, given user-defined workloads \textit{w}, objective function \textit{f} (i.e., minimizing any function of power, area, and latency/throughput), subject to cost constraints (e.g., maximum area \textit{a} or thermal design power \textit{p}). Our optimization problem may be described by:

\vspace{-1.5em}
\begin{align}
  \min_{h,w} f(h,w) \\
  \text{s.t.}\quad Area(h) \leq a\text{,} \quad TDP(h) \leq p\text{,} \\
  \quad ScheduleFailures(h, w) = 0,
\end{align}

The constraint $ScheduleFailures(h,w) = 0$ ensures that workload $w$ can be successfully mapped onto the architecture described by the hyperparameters $h$.

\subsection{FAST Framework Overview}
\label{sec:fastOverview}

The FAST framework explores the hardware datapath configuration, software schedule, and compiler operations for a combined search space up to $\mathcal{O}(10^{2300})$.
This estimate takes the product of the fully unconstrained mapspace \cite{timeloop} sizes for each layer in a moderately sized model like ResNet-50 (\mytilde$10^{2000}$), combined with the 10$^{13}$ datapath and 10$^{300}$ op fusion search spaces, rounded down. 

As shown in Figure \ref{fig:archSearch}, the FAST framework uses a three-phase approach for each trial.
Firstly, FAST uses Google Vizier \cite{vizier}, a black-box optimizer, to propose new choices of hyperparameters that define candidate hardware datapaths.
To make exploring the schedule space more tractable, Vizier constrains the software schedule mapspace to known-good mapping schemes such as weight and output-stationary \cite{Eyeriss}.
Secondly, our architectural simulator, described in Section \ref{sec:simulator},  simulates the mapping and execution of target workloads on the candidate architecture.
Compute-intensive ops such as Conv2D are optimized via pre-processing passes, such as tensor padding optimization, before calling Timeloop \cite{timeloop} with Vizier-provided constraints to determine the best schedule and predicted op performance. 
Finally, the per-op performance statistics are passed to our FAST fusion ILP solver to determine the best op fusion configuration. Our simulator then estimates op post-fusion performance and outputs final execution time and power for the target workloads.
This cycle then repeats for thousands of trials until convergence.
\looseness=-1

\subsection{Architectural Search Parameters}
\label{sec:params}

\begin{table} 
\vspace{0.5em}
\centering
\resizebox{0.43\textwidth}{!}{%
\begin{tabular}{| l | l | l |}
  \hline
  \textbf{Parameter Name} & \textbf{Type} & \textbf{Potential Values}\\
  \hline
  PEs\_x\_dim & int & 1 to 256, powers of 2\\
  \hline
  PEs\_y\_dim & int & 1 to 256, powers of 2\\
  \hline
  Systolic\_array\_x & int & 1 to 256, powers of 2\\
  \hline
  Systolic\_array\_y & int & 1 to 256, powers of 2\\
  \hline
  Vector\_unit\_multiplier & int & 1 to 16, powers of 2\\
  \hline
  L1\_buffer\_config & enum & Private, Shared\\
  \hline
  L1\_input\_buffer\_size & int & 1KB to 1MB, powers of 2\\
  \hline
  L1\_weight\_buffer\_size & int & 1KB to 1MB, powers of 2\\
  \hline
  L1\_output\_buffer\_size & int & 1KB to 1MB, powers of 2\\
  \hline
  L2\_buffer\_config & enum & Disabled, Private, Shared\\
  \hline
  L2\_input\_buffer\_multiplier & int & 1x to 128x, powers of 2\\
  \hline
  L2\_weight\_buffer\_multiplier & int & 1x to 128x, powers of 2\\
  \hline
  L2\_output\_buffer\_multiplier & int & 1x to 128x, powers of 2\\
  \hline
  L3\_global\_buffer\_size & int & 0MB to 256MB, powers of 2\\
  \hline
  GDDR6\_channels & int & 1 to 8, powers of 2\\
  \hline
  Native\_batch\_size & int & 1 to 256, powers of 2\\
  \hline
\end{tabular}}
\vspace{0.4em}
\caption{Accelerator datapath search space with $10^{13}$ possible values. When combined with scheduling and op fusion search spaces, the FAST total search space exceeds $10^{2300}$. Memory technologies besides GDDR6 can easily be modeled.}
\label{table:params}
\vspace*{-2.3em}
\end{table}

\begin{figure}[]
\centering
\vspace{-0.0em}
\includesvg[width=0.34\textwidth, keepaspectratio]{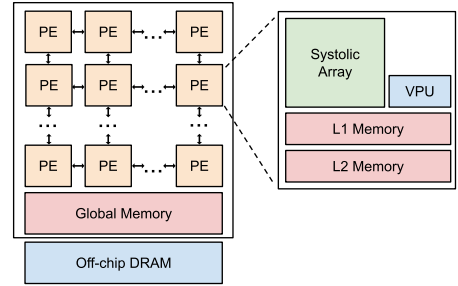}
\vspace{-0.4em}
\caption{Accelerator datapath configuration. PEs are connected by a mesh on-chip network. PE systolic arrays perform a matrix-vector multiply each cycle. Vector and scalar MACs can be modeled by setting systolic array X and/or Y dims to 1. PEs also contain a VPU for non-MAC vector ops. L2 and Global Memory structures are optional.} 
\label{fig:tpugrid}
\vspace{-1.4em}
\end{figure}

As shown in Figure \ref{fig:tpugrid}, we target a highly-parameterized and general ML accelerator template capable of modeling a wide range of previously proposed architectural designs.
Unlike prior work which targets specific families of accelerators, we enlarged our datapath search space to cover an approximate superset of popular accelerator families based on grids of processing elements (PEs), as described in Table \ref{table:params}.
In addition, to run models like BERT which require high performance non-MAC operations for layers such as softmax and layernorm, we add a fully general Vector Processing Unit (VPU) similar to TPU-v3 \cite{tpuv3}. VPU width, as a multiple of systolic array width, is added to our search space. 
The TPU family of accelerators instantiate large systolic arrays coupled with two levels of shared memory.
This can be represented in our framework by setting the systolic array dimensions to the appropriate values, setting L1\_buffer\_config to Shared, and L2\_buffer\_config to Disabled. 
Many accelerators such as Eyeriss \cite{Eyeriss} use flexible scalar PEs with per-PE buffers for input activations, weights, and output activations. 
This design can be reached by setting systolic array X and Y dimensions to 1, and L1\_buffer\_config to Private.
Several edge accelerators proposed in industry such as Simba \cite{simba} and EdgeTPU \cite{nahas} use vector PEs, which can be represented by setting the systolic array X dimension to 1.
While our datapath search space cannot perfectly cover all possible designs, it is still much larger than those used in previous work \cite{interstellar, magnet, miladhas}. 
We plan to further extend the search space in future work.
\looseness=-1

\subsection{FAST Fusion}
\label{sec:fusion}

Modern neural network models pose a challenge due to their poor operational intensity, as discussed in Section \ref{sec:opIntensityAndFusion}. 
We propose \textit{FAST fusion}, an aggressive fusion technique designed specifically for inference. 
FAST fusion leverages leftover Global Memory capacity unused by Timeloop to store activations and weights.
Activations have short lifetimes, typically until the next op. 
Weights can be stored to reuse across multiple inference requests, a technique called \textit{weight pinning}.
FAST fusion must balance the benefits of fusing activations with weights while focusing on memory-bound ops; there is typically no performance benefit for fusing a compute-bound op when latency can be hidden.
We express this constrained optimization problem as an \textit{integer linear program} (ILP) minimizing cycle count using simulator performance metrics. 
We implemented FAST fusion as a secondary pass that fuses XLA-generated HLO fusion regions. 
This improves ease of implementation and greatly reduces ILP problem size, at the cost of potentially suboptimal fusions.
FAST fusion conservatively assumes that entire tensors are stored in memory; schedulers can use inter-op blocking to reduce tensor working set sizes.
\looseness=-1

Next, we describe the problem formulation. We are given an input graph $G(V,E)$ representing an $n-$layer, partially fused\footnote{That is, $G(V,E)$ is derived from an original $m-$layer network that has been optimized such that combinations of data formatting, element-wise, and matrix operations have been grouped in fused computations~\cite{tpuv3}.} neural network which we wish to optimize, where each vertex $v \in V$ is a layer of the network, while each edge $e = (u,v)$ represents an activation dependency from layer $u$ to $v$ (that is, the output activation of $u$ is an input to $v$). Let $F_{in}(v)$ and $F_{out}(v)$ represent the fan-in and fan-out sets, respectively, of some vertex $v \in V$. We assume that $G$ has the property that while $0 \leq \vert F_o(v) \vert \leq n-1$, $0 \leq \vert F_i(v) \vert \leq 1$. To simplify notation, let $D_t := \{I, O, W\}$ represent the set of data types used to annotate variables, where $I$, $O$, and $W$ represent input activations, output activations, and weights, respectively. 
Given a known execution order $o:~v \in V \longrightarrow \{0,\ldots,n-1\}$ for each network layer, we express the optimization problem in Figure \ref{fig:optimization}.
\looseness=-1

\begin{figure}

\vspace*{0.4em}
\begin{mini}
  {p_i^k}{\sum_{i \in V} T_i}{}{}
  \addConstraint{T_i}{\geq T_i^{min}}{}
  \addConstraint{T_i}{\geq T_i^{max} - \sum_{k \in D_t} t_i^k \cdot p_i^k}{}
  \addConstraint{C_{GM}}{\geq B_i + \sum_{k \in D_t} d_i^k \cdot p_i^k + \sum_{j \in V, j \neq i}W_j \cdot p_j^W}{}
  \addConstraint{p_i^O}{\geq p_j^I}{~~~~~~~~~~~~\forall j \in F_{out}(i)}
  \addConstraint{\sum_{j \in F_{out}(i)}p_j^I}{\geq p_i^O}{}
  \addConstraint{M \cdot (1 - p_i^I)}{\geq o(i) - o(F_{in}(i)) - 1}{}
  \addConstraint{p_i^k}{\in \{ 0, 1 \}}{~~~~~~~~~~~~\forall k \in D_t}
\end{mini}
\vspace{-1.0em}
\caption{Optimization problem for FAST fusion.}
\vspace{-1.4em}
\label{fig:optimization}
\end{figure}

The variable $p_i^k$ is a binary decision variable indicating whether the tensor of type $k \in D_t$ for layer $i$ is to be placed in the Global Memory (if equal to 1), while the variable $T_i$ represents the optimized execution time for layer $i$ as a function of $p_i^k$.~$T_i^{min}$ and $T_i^{max}$ are the execution times for layer $i$ when the inputs and outputs of the layer are pinned exclusively in the Global Memory and DRAM, respectively (these are obtained from Timeloop evaluation of the layer). The parameter $t_i^k$ is time to access layer $i$'s tensor of type $k$ (where $k \in D_t$) from DRAM, $C_{GM}$ is the capacity of the Global Memory in bytes, $B_i$ is the nominal global buffer usage of layer $i$, $d_i^k$ is the difference between the size of layer $i$'s tensor of type $k$ and the corresponding tile size allocated on the global buffer if we were to assume the tensor is being streamed from/to DRAM, $W_j$ is the size of layer $j$'s weight tensor, and $M \geq n - 1$ is an arbitrarily large constant. Note that the constraints imply that activations are only stored in the global buffer if the op consuming an activation executes immediately after the op which produces the activation. This also means that in cases where a node has multi-fanout (e.g., skip connections), at most only one op in the fanout cone will benefit from reading its input activation from global memory. These constraints -- which limit the maximum potential upside of the technique -- were imposed because of some limitations in our simulation infrastructure. Future work will address these limitations, thereby potentially allowing for further performance gains.
\looseness=-1



\subsection{Two-Pass Softmax}
\label{sec:softmax}

The softmax operation is challenging on the TPU-v3 and other existing ASICs for several reasons. Firstly, calculating an exponential operation requires significant hardware resources. Typically, a look-up table is used with a Taylor series expansion, resulting in a large latency and area \cite{exponential}. Secondly, numerically-stable softmax requires 3 passes over the input vector, as shown in Algorithm~\ref{alg:softmax}. Due to the size of the vector in most models, these 3 passes usually involve reading and writing the values to and from DRAM.

In order to reduce the number of memory accesses, \cite{twopasssoftmax} proposes a mathematically equivalent algorithm that performs the first 2 passes together, as seen in Algorithm~\ref{alg:twopasssoftmax}.
The two-pass softmax eliminates $N$ memory accesses relative to Algorithm~\ref{alg:softmax}, but increases the number of exponential calculations by up to $2N$.
Therefore, the benefit of the two-pass approach is dependent on the accelerator's memory bandwidth and vector unit throughput. We add the option to enable the two-pass softmax as a hyperparameter within our FAST search space.
\looseness=-1

\begin{figure}
\vspace{-0.5em}
\begin{algorithm}[H]
{\fontsize{8}{8}\selectfont
\caption{Numerically-Stable Softmax}\label{alg:softmax}
  \begin{algorithmic}[1]
  
\State $maxVal \gets -\infty$
\For{$i \gets 1$ to $N$}
    \State $maxVal \gets max(maxVal, V[i])$
\EndFor

\State $sum \gets 0$
\For{$i \gets 1$ to $N$}
    \State $tempVec[i] \gets exp(V[i] - maxVal)$
    \State $sum \gets sum + tempVec[i]$
\EndFor

\For{$i \gets 1$ to $N$}
    \State $out[i] \gets tempVec[i] / sum $
\EndFor

\end{algorithmic}
}
\end{algorithm}
\vspace{-1.5em}

\begin{algorithm}[H]
{\fontsize{8}{8}\selectfont

\caption{Two-Pass Softmax}\label{alg:twopasssoftmax}
  \begin{algorithmic}[1]
\State $runningMax \gets -\infty$
\State $runningSum \gets 0$
\For{$i \gets 1$ to $N$}
    \State $newMax \gets max(runningMax, V[i])$
    \State $runningSum \gets runningSum * exp(runningMax-newMax) + exp(V[i] - newMax)$
    \State $runningMax \gets newMax$
\EndFor

\For{$i \gets 1$ to $N$}
    \State $out[i] \gets exp(V[i]) / runningSum $
\EndFor

\end{algorithmic}
}
\end{algorithm}
\vspace{-2.0em}
\end{figure}

%% file: results.tex
\section{Evaluation}
\label{sec:evaluation}

\begin{figure*}[t]
\centering
\vspace{-0.2em}
\includesvg[width=1.0\textwidth, keepaspectratio]{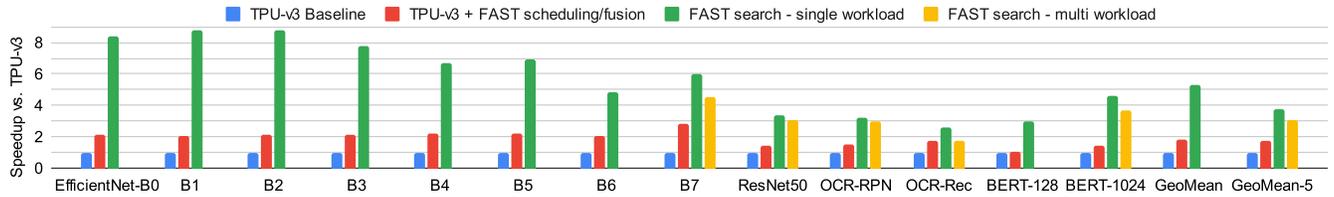}
\vspace{-2.0em}
\caption{Modeled inference throughput relative to TPU-v3. "FAST scheduling/fusion" optimization offers large speedups over existing TPU-v3. Speedups are much larger when FAST is also allowed to search over the datapath. "FAST search - single workload" are optimized designs for a specific workload. "FAST-search - multi workload" is a single design optimized across the 5 workloads (i.e., EfficientNet-B7, ResNet50, OCR-RPN, OCR-Rec, BERT-1024). GeoMean and GeoMean-5 results correspond to the geometric mean across all workloads, and across the aforementioned 5 workloads respectively.} 
\label{fig:speedupPerf}
\vspace{-0.7em}
\end{figure*}

\begin{figure*}[]
\centering
\includesvg[width=1.0\textwidth, keepaspectratio]{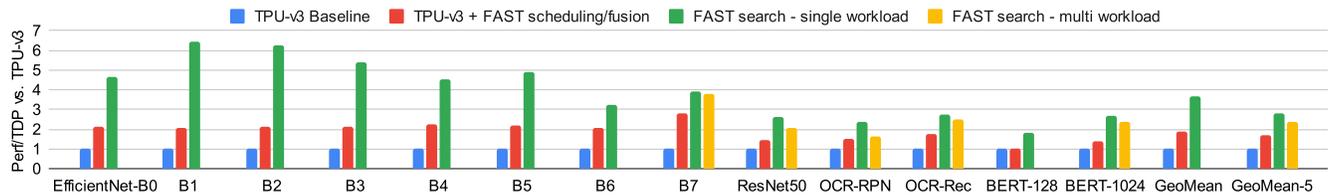}
\vspace{-2.0em}
\caption{Modeled inference throughput per TDP (peak power draw) relative to TPU-v3, normalized to the same manufacturing process technology. FAST demonstrates large Perf/TDP wins across all workloads.} 
\label{fig:speedupPerfTDP}
\vspace{-1.5em}
\end{figure*}

\subsection{Experimental Setup}
\noindent \textbf{Methodology and Simulator:}
\label{sec:simulator}
We use TPUs as the baseline because TPUs are well-characterized as the most popular dedicated datacenter ML accelerator; TPUs vs GPUs have also been well-studied in prior work \cite{tpupaper2017} \cite{tpuv4i} \cite{mlperf} \cite{efficientnettpu}. We modified an internal TPU performance simulator to enable modeling of a wide range of architectures as described in Section \ref{sec:params}.
The baseline simulator is well-correlated: on our benchmark suite, simulator accuracy is on average within $8.2\pm2.7\%$ of profiled TPU-v3 performance.
Because our simulator tends to produce slightly optimistic results, we evaluated against a simulated rather than measured TPUv3 baseline to take optimistic simulator assumptions into account.
\looseness=-1

Our simulator takes unmodified XLA HLO graphs \cite{xla} as input and is modified to employ Timeloop \cite{timeloop} to evaluate the performance of Conv2D, DepthwiseConv2D, Einsum, and MatMul operations. 
Since Timeloop cannot handle problem dimensions that do not factorize cleanly into hardware datapath dimensions, we added a padding preprocessing step to improve utilization.
All other ops, such as vector ops used in softmax, are modeled using our simulator's custom cost models. 
Simulated design points with Timeloop scheduling failures are considered invalid. 
To estimate area and power consumption, we built analytical models correlated to production designs on an industry sub-10-nm manufacturing process; for a fair comparison, the TPU-v3 baseline is also modeled using this same process technology.
TDP is estimated as \textit{power virus power}, in which each component is assumed to be accessed at 100\% utilization.
FAST fusion's ILP is solved using \texttt{SCIP v7.0.1}~\cite{GamrathEtal2020OO}, and is configured with a 20 minute time-out; if an optimal solution is not found in that time the solver returns the best incumbent solution.
Because we use pre-fused XLA HLO graphs as input, our FAST fusion implementation fuses XLA-generated fusion regions instead of individual ops. 
\looseness=-1

\noindent \textbf{Workloads:} 
FAST is evaluated on a range of state-of-the-art workloads in both computer vision and natural language processing domains.
The entire suite of EfficientNets is evaluated, from B0 to B7 \cite{efficientnet}.
BERT-Base \cite{bert} is evaluated for both short (128) and long (1024) sequence lengths. 
ResNet50v2 \cite{resnet50} is one of the most popular CNN-based models.
We also evaluated two components of a production OCR pipeline described in \cite{vssOcr}.
OCR-RPN is the first stage in a standard Mask R-CNN implementation used to propose candidate text regions of interest.
OCR-Recognizer is an LSTM-based model within the recognizer pipeline.
These workloads were selected based on their range of performance characteristics on TPU-v3.
EfficientNets currently run less efficiently on TPUs due to their use of depthwise-separable convolutions (see Section \ref{sec:efficientnetPerf}).
BERT runs efficiently on TPUs at short sequence lengths, but is less efficient at longer sequence lengths (see Section \ref{sec:bertPerf}); we capture both by evaluating BERT with sequence lengths 128 and 1024.
ResNet50v2 runs much more efficiently than EfficientNet by using standard Conv2D operations.
OCR-RPN and OCR-Recognizer are already optimized to run efficiently on TPUs, and represent a worst-case scenario: 
models that already run efficiently on our TPU-v3 baseline will benefit less from FAST. 
\looseness=-1

\noindent \textbf{Optimization framework:}
We used Google Vizier \cite{vizier} enabling LCS optimization \cite{hyperlcs} and safe search \cite{bayesian-gelbert2014}, disabling transfer learning, with 5000 trials per experiment. Each trial takes 10 minutes to 2 hours wall clock time based on model size and datapath constraints.

\subsection{Experimental Results}\label{sec:results}

\subsubsection{Overall Speedup}

Figure \ref{fig:speedupPerf} shows overall performance improvement from FAST-generated custom accelerators on each workload relative to a simulated TPU-v3 baseline, in which performance is measured in processed inference queries per second (QPS).
FAST is given a power and area budget similar to the current-generation TPU-v3, but on a new process technology, emulating the methodology used by accelerator architects to design next-generation products. 
The purpose of this experiment is to evaluate if FAST can fully utilize the available power and area headroom to create high-performance designs. 
We evaluate FAST optimizing for individual workloads as well as across multiple workloads. 
Our multi-workload optimized FAST finds a single hardware design evaluated on the geometric mean across EfficientNet-B7, ResNet50v2, OCR-RPN, OCR-Recognizer, and BERT-1024 achieving a 3.1$\times$ speedup over TPU-v3 baseline. 
Overall speedups are much higher on EfficientNets due to their use of depthwise separable convolutions.
When provided with pure performance as the objective, FAST successfully finds large designs that come close to our maximum area and TDP constraints.
OCR-RPN and OCR-Recognizer are already well-optimized for TPU-v3 and thus have the lowest gains as expected.
Utilizing FAST-specified scheduling and fusion on the TPU-v3 datapath provides a substantial 1.7$\times$ speedup; however, this is optimistic since implementing the generated schedules and achieving the projected speedup may require hardware changes. 
Tuning an architecture across multiple workloads results in slightly reduced, but still substantial improvements over the baseline. 
FAST search achieves a 3.8$\times$ average speedup when optimizing for single workloads.
\looseness=-1

Absolute performance numbers can be misleading since different hardware designs vary in cost. 
A common metric for evaluating datacenter accelerators is to normalize for these differences by considering performance per TCO, which includes initial capital expenses and recurring operating costs such as electricity (Section \ref{sec:roi}). 
While due to sensitivity of data, TCO numbers are not published, TDP can be used as a proxy for TCO \cite{tpupaper2017}.
Figure \ref{fig:speedupPerfTDP} shows Perf/TDP numbers relative to a hypothetical TPU-v3 die-shrunk to the same sub-10-nm process technology. 
When optimizing for Perf/TDP, FAST finds balanced designs that are smaller than our maximum area and TDP constraints, but achieve high compute utilization with minimal memory bandwidth bottlenecks.
Designs found by FAST tend to have smaller systolic arrays, smaller L1 scratchpads, and larger Global Memories than the TPU-v3 baseline.
Two-pass softmax was not useful when fusion was enabled.
Overall, FAST individually optimized for each workload improves Perf/TDP on average by 3.7$\times$ across all workloads and 2.8$\times$ on the reduced workload suite, whereas FAST optimized for multiple workloads still improves Perf/TDP by 2.4$\times$.

\subsubsection{ROI Discussion}

\begin{table} 
\centering
\vspace{0.8em}
\resizebox{0.46\textwidth}{!}{%
\begin{tabular}{| l | l | l | l | l | l |}
  \hline
   \textbf{Target Workload} & \textbf{Perf/TCO} & \textbf{1x ROI} & \textbf{2x ROI} & \textbf{4x ROI} & \textbf{8x ROI}\\
  \hline
  EfficientNet-B7 & 3.91x & 2,164  & 4,327 & 8,655 & 17,309\\
  \hline
  ResNet50 & 2.65x & 2,588 & 5,177 & 10,354 & 20,707 \\
  \hline
  OCR-RPN & 2.34x & 2,810 & 5,620 & 11,241 & 22,482 \\
  \hline
  OCR-Rec & 2.72x & 2,548 & 5,096 & 10,192 & 20,385 \\
  \hline
  BERT-128 & 1.84x & 3,534 & 7,069 & 14,138 & 28,276 \\
  \hline
  BERT-1024 & 2.7x & 2,558 & 5,115 & 10,231 & 20,462 \\
  \hline
  Multi-Workload & 2.82x & 2,792 & 5,584 & 11,167 & 22,335 \\
  \hline
\end{tabular}}
\vspace{0.3em}
\caption{FAST-generated accelerator deployment volume required to reach a specific ROI target based on Perf/TDP speedups from Figure \ref{fig:speedupPerfTDP}. An ROI above 1 is profitable. "Multi-Workload" is optimized across a set of workloads: EfficientNet-B7, ResNet50, OCR-RPN, OCR-Rec, BERT-1024.}
\label{table:roiCrossover}
\vspace{-2.3em}
\end{table}

To determine the practicality of building FAST-generated ML accelerators optimized for single workloads, we estimate FAST accelerator ROI relative to our TPUv3 baseline in Table \ref{table:roiCrossover}, using the methodology and parameters as described in Section \ref{sec:roi}.
We estimate Perf/TCO speedup relative to TPUv3 using the Perf/TDP speedups shown in Figure \ref{fig:speedupPerfTDP}. Prior work shows that Perf/TCO and Perf/TDP are highly correlated \cite{tpuv4i}.
TPUs are not available for sale to the general public and TPU TCO is confidential; we instead used the NVIDIA DGX A100 320GB platform to approximate TPUv3 TCO.
This example is intended as an estimate for illustrative purposes only; entities planning to build their own custom accelerators should calculate ROI using costs and baselines specific to their own situation.

Our analysis shows that the ROI 1x break-even point can be reached with a deployment volume of 2,164 to 3,534 accelerators.
However, breaking even is typically not enough to justify building custom accelerators; the goal is typically to turn a profit. 
Many corporations make business decisions based on whether a planned project reaches a pre-determined ROI threshold \cite{roi}.
Due to ROI diminishing returns from increasing Perf/TCO, it is typically better to target larger deployment volumes, thus suggesting that the FAST-generated design optimized for 5 different workloads (Multi-Workload) may be a more profitable design since it can likely be deployed in a much larger volume without much impact on Perf/TCO. 
Even single-workload designs can easily reach profitability with typical deployment sizes discussed in Section \ref{sec:roi}.

\subsubsection{Search Convergence Rate}

\begin{figure}[]
\centering
\vspace{-0.0em}
\includesvg[width=0.43\textwidth, keepaspectratio]{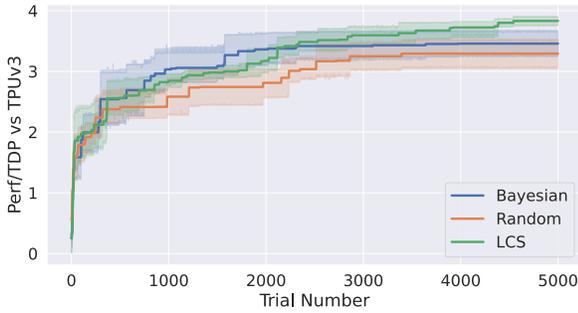}
\vspace{-1.2em}
\caption{Search convergence rate on EfficientNet-B7 for Bayesian, random, and Linear Combination Swarm \cite{hyperlcs}.} 
\label{fig:convergenceRate}
\vspace{-0.7em}
\end{figure}

We evaluated several black box optimizer heuristics as provided by Vizier. 
In Figure \ref{fig:convergenceRate}, we compare the convergence rate of Vizier's default Bayesian algorithm against Linear Combination Swarm (LCS) \cite{hyperlcs} and random sampling when optimizing for Perf/TDP on EfficientNet-B7.
We show the mean and 90\% confidence interval across each heuristic, across 5 runs per heuristic.
LCS outperforms the other heuristics when trials exceed 2000.
\looseness=-1

\subsubsection{Pareto Frontier}

\begin{figure}[]
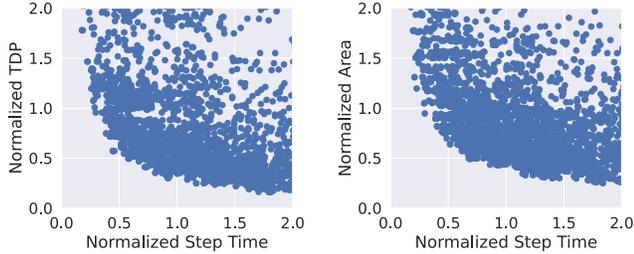

\centering
\vspace{-0.2em}
\hspace{-0.5em}
\subfloat{{
\includesvg[width=0.221\textwidth, keepaspectratio]{figures/tdp_pareto.svg}
}}
\quad
\subfloat{{
\includesvg[width=0.221\textwidth, keepaspectratio]{figures/area_pareto.svg}
}}
\vspace{-0.7em}
\caption{EfficientNet-B7 step time vs. TDP and area relative to TPU-v3 on the same process technology.} 
\label{fig:pareto}
\vspace{-0.7em}
\end{figure}

To evaluate our search space coverage, in Figure \ref{fig:pareto} we characterize the relationship between performance, TDP and area on EfficientNet-B7.
Each figure is normalized to a hypothetical TPU-v3 built with the same sub-10nm process technology at (1.0, 1.0), and points located towards the lower left are Pareto-optimal.
FAST is able to find a range of designs significantly better than the baseline, suggesting that FAST can be easily applied to other domains besides datacenters, such as targeting embedded systems.
\looseness=-1

\subsubsection{Example Designs Found by FAST}

\begin{table} 
\centering
\hspace*{-0.1em}
\vspace{0.0em}
\resizebox{0.47\textwidth}{!}{%
\begin{tabular}{| l | l | l | l |}
  \hline
   & \textbf{Modeled TPU-v3} & \textbf{FAST-Large} & \textbf{FAST-Small}\\
  \hline
  Normalized TDP  & 0.5x  & 0.4x  & 0.15x\\
  \hline
  Normalized Area & 0.6x  & 0.7x  & 0.3x\\
  \hline
  Peak Compute    & 123 TFLOPS & 131 TFLOPS & 32 TFLOPS \\
  \hline
  Peak Bandwidth & 900 GB/s & 448 GB/s & 448 GB/s\\
  \hline
  Batch Size & 2x64 & 8 & 64\\
  \hline
  Num PEs & 2x2 & 64 & 8\\
  \hline
  PE Systolic Array Dims & 128x128 & 32x32 & 64x32\\
  \hline
  PE Vector Width & 512 & 32 & 64\\
  \hline
  PE L1 Buffer Size & 2x64KiB & 8 KiB & 8 KiB\\
  \hline
  PE L1 Buffer Config & Shared & Shared & Shared\\
  \hline
  PE L2 Buffer Config & Disabled & Disabled & Disabled\\
  \hline
  Global Buffer Size & 2x16 MiB & 128 MiB & 8 MiB\\
  \hline
  Compute Utilization & 0.14 & 0.61 & 0.74\\
  \hline
  Pre-fusion Mem Stall \% &   & 63\% & 21\%\\
  \hline
  Fusion Efficiency &  & 85\% & 0\%\\
  \hline
  OpInt Ridgepoint & 137 & 292 & 73\\
  \hline
  Fused Model OpInt & 63 & 383 & 63\\
  \hline
  B7 Performance (QPS) & 210 (aggregate) & 733 & 241\\
  \hline
  B7 Inference Latency & 609ms & 11ms & 265ms\\
  \hline
  Normalized Perf/TDP & 1 & 3.9 & 3.9\\
  \hline
\end{tabular}}
\vspace{0.3em}
\caption{Two example designs found by FAST optimized for EfficientNet-B7 with similar overall Perf/TDP. Area and power are normalized to threshold constraints.}
\label{table:fastDesigns}
\vspace{-2.7em}
\end{table}

Table \ref{table:fastDesigns} shows two example designs found with FAST when optimizing Perf/TDP on EfficientNet-B7, compared to TPU-v3 normalized to the same sub-10nm process technology.
TPU-v3 is a dual-core design, in which each core is treated as a separate accelerator; our EfficientNet-B7 QPS results show aggregate QPS when using both cores.
Each TPU-v3 core contains two 128x128 systolic arrays and a 1024-wide vector unit. 
FAST-generated designs are all single-core, in which each core contains one or more PEs.
EfficientNet-B7 when executed on TPU-v3 is both compute-bound (low compute utilization of 0.14) and memory-bound (low operational intensity of 63), both of which are addressed by FAST designs with different approaches. 
Overall, FAST preferred small shared L1 buffers with no L2 buffer; although L2 buffers may reduce dynamic power from improved blocking, they increase overall TDP when assuming maximum buffer accesses per cycle. 
To improve mapping efficiency for depthwise-separable convolutions, both designs have PEs with smaller systolic array dimensions resulting in significantly higher compute utilization. Despite FAST-Large having similar peak compute performance as TPU-v3 with half the peak memory bandwidth, the design is not bandwidth-bottlenecked due to its 128MiB Global Buffer, enabling aggressive FAST fusion to improve operational intensity from 63 to 383. Overall, idle time spent waiting for DRAM transfers to complete is reduced by 85\%, from 63\% pre-fusion to 9\% post-fusion. FAST-Small avoids fusion entirely and achieves high efficiency through a low compute to memory bandwidth ratio.
Although both designs have similar Perf/TDP, FAST-Large is preferred for datacenter environments because it meets MLPerf image classification latency requirements (15ms) \cite{mlperfInfrules}, enabling EfficientNet-B7 for latency-sensitive applications.

\subsubsection{Evaluating FAST Fusion}
\begin{figure}[]
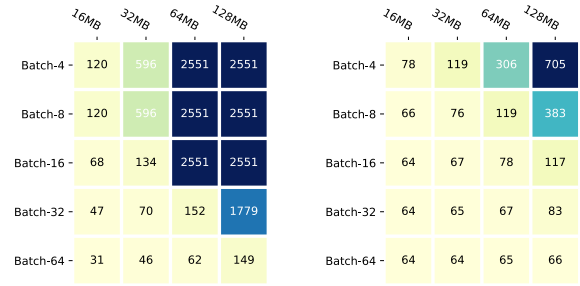

\captionsetup[subfloat]{oneside,margin={0.0cm,0cm}}
\subfloat[EfficientNet-B0]{{
\includesvg[width=0.19\textwidth, keepaspectratio]{figures/opIntHeatmap_b0.svg}
}}
\qquad
\subfloat[EfficientNet-B7]{{
\includesvg[width=0.19\textwidth, keepaspectratio]{figures/opIntHeatmap_b7.svg}
}}
\vspace{-0.8em}
\caption{FAST-Large post-fusion operational intensity, sweeping Global Memory (columns) and batch size (rows). Operational intensity increases with larger Global Memory and smaller batch size. Higher is better, but there is no more performance benefit after reaching the ridgepoint (292).} 
\label{fig:opIntHeatmap}
\vspace{-1.5em}
\end{figure}

We evaluate FAST fusion performance in Figure \ref{fig:opIntHeatmap} by measuring its impact on operational intensity as we sweep Global Memory and batch size in an otherwise-fixed FAST-Large design.
Increasing Global Memory capacity enables FAST fusion to assign more activation and weight tensors from DRAM to Global Memory, resulting in higher operational intensity.
Decreasing batch size decreases tensor activation size (see Table \ref{table:efficientNetSizes}), increasing operational intensity since more tensors can be kept in Global Memory.
However, decreasing batch size also potentially reduces compute utilization due to decreased parallelism.
Therefore, the goal of FAST fusion is to select the largest batch size in which post-fusion operational intensity meets or exceeds the accelerator ridgepoint (292 for FAST-Large).
EfficientNet-B0 has small activation and weight tensor sizes, making it easy for FAST fusion to exceed 292.
The largest EfficientNet model, B7, represents a worst-case scenario for fusion, but FAST fusion can still achieve sufficiently high operational intensity at batch size 8 to overcome the memory bottleneck.

\subsubsection{Performance Characterization}

\begin{figure}[]
\centering
\vspace{-0.0em}
\includesvg[width=0.48\textwidth, keepaspectratio]{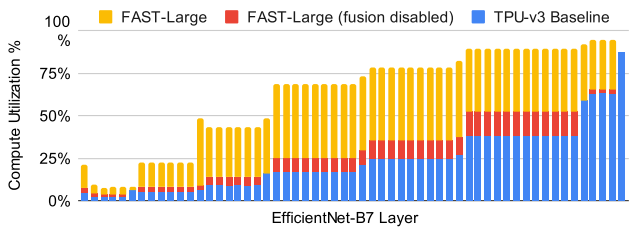}
\vspace{-2.4em}
\caption{FAST-Large EfficientNet-B7 per-layer performance as a fraction of peak FLOPS. Changing from TPU-v3's 128x128 systolic arrays to FAST-Large's 32x32 systolic arrays improves compute utilization, but remains bottlenecked by memory bandwidth until FAST fusion is enabled.} 
\label{fig:perlayerperfAfter}
\vspace{-0.7em}
\end{figure}

\label{sec:perfBreakdown}

\begin{figure}[]
\centering
\vspace{-0.0em}
\hspace*{-0.7em}
\includesvg[width=0.5\textwidth, keepaspectratio]{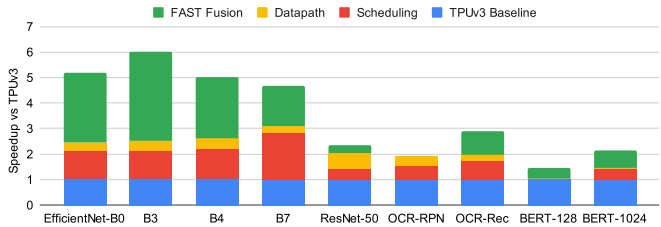}
\vspace{-2.4em}
\caption{Performance breakdown of each component of FAST relative to a modeled TPU-v3 single core baseline. Improvements are additive; for example, FAST fusion includes both datapath and scheduling improvements.} 
\label{fig:perfBreakdown}
\vspace{-1.2em}
\end{figure}

In Figure \ref{fig:perfBreakdown}, we show the contributions of improved scheduling from Timeloop, datapath improvements, and FAST fusion.
We start with the TPU-v3 baseline, and then incrementally add improvements from FAST-Large. 
Since TPU-v3 is a dual-core design whereas FAST-Large is single-core, we compare a single TPU-v3 core against a halved FAST-Large design with 32 PEs.
The scheduling component shows the potential speedup from better mappings discovered by Timeloop; implementing these better mappings may require changes to the hardware. 
In the datapath component, the TPU-v3's 128x128 systolic arrays are replaced with 32x32 systolic arrays, keeping peak FLOPS constant. We also replace TPU-v3's 16MB global memory with the FAST-discovered 128MB global memory.
Datapath improvements without FAST fusion result in significantly lower speedups since performance is a function of both compute and memory, and increasing compute utilization results in no further improvements once the memory bandwidth limit is reached. 
There is no performance benefit from increasing global memory size when fusion is disabled.
Enabling FAST fusion removes the memory bandwidth bottleneck, allowing the improved datapath to realize its utilization improvements. 
Scheduling, datapath, and fusion all work in synergy to achieve FAST's projected speedups, thereby demonstrating the criticality of including fusion when performing hardware datapath + scheduling co-optimization to address both memory and compute bottlenecks.
\looseness=-1

\begin{table} 
\centering
\vspace{0.2em}
\resizebox{0.47\textwidth}{!}{%
\begin{tabular}{| l | l | l | l |}
  \hline
   & \textbf{EfficientNet-B7} & \textbf{ResNet50} & \textbf{BERT-Seq1024}\\
  \hline
  FAST-Large & 4.27x (1.00)  & 2.95x (1.00) & 2.39x (1.00)\\
  \hline
  With 16MB Global Mem & 2.26x (0.53) & 2.20x (0.75) & 1.22x (0.51) \\
  \hline
  Without FAST Fusion & 1.91x (0.45) & 1.74x (0.59) & 1.05x (0.44) \\
  \hline
  With 128x128 systolic arrays & 2.69x (0.63) & 1.41x (0.48) & 1.35x (0.56)\\
  \hline
  With 32KB L1 scratchpads & 3.20x (0.75) & 2.26x (0.77) & 1.83x (0.77)\\
  \hline
\end{tabular}}
\vspace{0.3em}
\caption{FAST-Large ablation study measuring Perf/TDP relative to a die-shrunk TPUv3 baseline. Numbers in parentheses show Perf/TDP relative to FAST-Large. The first row shows an unmodified FAST-Large baseline. Subsequent rows show FAST-Large with a single component reverted to the TPUv3 baseline with no other changes.
\looseness=-1}
\label{table:ablation}
\vspace{-2.2em}
\end{table}

In Table \ref{table:ablation}, we performed an \textit{ablation study} to evaluate FAST-Large performance relative to TPU-v3. 
An ablation study characterizes system performance by removing individual components to understand each component's contribution to the overall system.
The first row shows unmodified FAST-Large Perf/TDP relative to our die-shrunk TPU-v3 baseline. Subsequent rows evaluate FAST-Large with a single component replaced with what was used in the original TPU-v3 design with no other changes. For example, the second row evaluates FAST-Large with its 128MB Global Memory replaced with the 16MB Global Memory used in TPU-v3. The resulting performance loss of each row relative to the first row represents the feature's importance towards achieving good overall performance.
Moving from 32KB to 8KB L1 scratchpads has a minimal impact on performance with lowered power, resulting in improved Perf/TDP.
Removing each component optimization results in substantial Perf/TDP degradation, thereby demonstrating their importance for FAST-Large.

\section {Conclusion and Future Work}
We presented FAST, a full-stack accelerator search technique that performs joint optimization of the hardware datapath, software scheduling, and compiler passes such as operation fusion and tensor padding.
Our results demonstrate that FAST-generated inference accelerators can provide large Perf/TDP improvements on state-of-the-art computer vision and natural language processing models with compelling ROIs. For example, the FAST-Large design provides 3.9$\times$ Perf/TDP improvement over a die-shrunk TPU-v3 baseline through a combination of higher efficiency and lower optimal batch size, thus enabling not only substantial Perf/TCO improvements but also enables EfficientNet-B7 to be deployed for latency-sensitive applications. 
FAST-generated accelerators currently do not support the full feature set provided by designs like TPU-v3 optimized for not just single-chip inference, but also training across thousands of devices.
However, key inference datacenter workloads may be sufficiently important or provide sufficient volume for substantial returns on investment.
Specialized designs optimized for small sets of workloads are unlikely to completely replace general-purpose designs, but may still serve an important niche in production environments.
By substantially enlarging the set of workloads, FAST may also be used to propose the design of future generations of general-purpose ML accelerators.
We plan to extend FAST by further enlarging the search space and adding support for optimizing accelerators for training.